\documentclass[runningheads]{llncs}

\usepackage[T1]{fontenc}

\usepackage{graphicx}
\usepackage{floatrow, makecell, tabularx, array, booktabs} 
\usepackage[english]{babel}
\usepackage[colorlinks=true, allcolors=blue]{hyperref}
\usepackage{bigstrut}
\usepackage{multirow}
\usepackage{listings}

\usepackage{amsfonts}

\usepackage{pifont}

\newcommand{\xmark}{\ding{55}}

\usepackage[backend=bibtex,style=numeric, maxnames=3]{biblatex}
\begin{document}

\title{A request for clarity over the End of Sequence token in the Self-Critical Sequence Training}


\author{Jia Cheng Hu\inst{1}\orcidID{0009-0008-1611-966X} \and
Roberto Cavicchioli\inst{1}\orcidID{0000-0003-0166-0898} \and
Alessandro Capotondi\inst{1}\orcidID{0000-0001-8705-0761}}

\institute{University of Modena and Reggio Emilia \\ \email{name.surname@unimore.it}}

\authorrunning{J.C. Hu et al.}

\titlerunning{A request for clarity over the EoS token in the SCST}

\maketitle

\selectlanguage{english}

\begin{abstract}

\textit{The Image Captioning research field is currently compromised by the lack of transparency and awareness over the End-of-Sequence token (<Eos>) in the Self-Critical Sequence Training. If the <Eos> token is omitted, a model can boost its performance up to +4.1 CIDEr-D using trivial sentence fragments. While this phenomenon poses an obstacle to a fair evaluation and comparison of established works, people involved in new projects are given the arduous choice between lower scores and unsatisfactory descriptions due to the competitive nature of the research. This work proposes to solve the problem by spreading awareness of the issue itself. In particular, we invite future works to share a simple and informative signature with the help of a library called SacreEOS. Code available at \emph{\href{https://github.com/jchenghu/sacreeos}{https://github.com/jchenghu/sacreeos}}}. 

\end{abstract}

\section{Introduction}

The standard training strategy of a modern Neural Image Captioning system includes a policy gradient method, called Self-Critical Sequence Training \cite{rennie2017self} (shortened as SCST) which is designed to maximize the evaluation score given to the outputs. 
In this work, we discuss the problems caused by the lack of transparency from the research community over the inclusion or omission of the End-of-Sequence token during the optimization. An easy-to-overlook implementation detail that can significantly increase the performance of any model despite yielding worse descriptions. 

The lack of awareness of the impact of the End-of-Sequence (\texttt{<Eos>}) omission and the lack of explicit information on the SCST implementation during the reporting of results pose an obstacle to scientific progress as they make it challenging to compare established works and evaluate new ones. 
Our paper attempts to spread awareness about the issue and proposes a solution to increase transparency in future works. This paper is structured as follows: in Section \ref{sec:prob}, we discuss the problem of the End-of-Sequence omission and why it is a problem for the research community; in Section \ref{sec:ana}, we provide a qualitative and quantitative analysis of the issue and we sample some of the recent works in Image Captioning to demonstrate its pervasiveness and provide some practical examples of its impact; In Section \ref{sec:sacre}, we propose a possible solution with the help of a Python library called SacreEOS; in Section \ref{sec:rel}, we mention some of the literature approaches, and, finally, we draw our conclusions in Section \ref{sec:conc}.

\section{Problem Description}
\label{sec:prob}


\subsection{CIDEr Optimization}
CIDEr \cite{vedantam2015cider} is an n-gram-based metric that evaluates the caption semantic content according to its similarities to the ground truths. Compared to the other metrics \cite{papineni2002bleu, lin2004rouge, banerjee2005meteor, anderson2016spice}, it exploits the entire corpus of reference descriptions in the attempt of backing the evaluation with the consensus of the majority of people. In particular, each n-gram $w_k$ in sequence $Z$ is weighted according to the \textit{tf-idf} term $g^n_k(Z)$ defined as:
\begin{equation}
    \frac{h^n_k(Z)}{\sum\limits_{w_l \in \Omega} h^n_l(Z) } \cdot log (  \frac{|I|}{\sum\limits_{I_i \in I} \min(1, \sum\limits_{q} h^n_k(V^i_q)) } )
\label{idf_equation}
\end{equation}
where $\Omega$ is the set possible n-grams in the corpus, $I$ is the set of corpus images and $h^n_k(Z)$, $h^n_k(V^{i}_{j})$ represent the number of occurrences of n-gram $w_k$ in the sequence $Z$ and in the j-th ground truth of image $I_{i} \in I$. The CIDEr and its alternative (CIDEr-D), compute the similarity between the candidate and reference description as the number of matching n-grams, weighted according to Equation \ref{idf_equation}. We refer to \cite{vedantam2015cider} for additional details of the formula since they are unnecessary for the sake of the discussion. 

The standard training practice of the Image Captioning model consists of a pre-training phase using the Cross-Entropy loss followed by a CIDer-D optimization by means of a policy gradient method called Self-Critical Sequence Training \cite{rennie2017self}. The latter minimizes the negative expected reward:
\begin{equation}
    L_{R}(\theta)=-\mathbf{E}_{y_{1:T} \sim p_{\theta}}[r(y_{1:T})]
    \label{expected}
\end{equation}
where $r$ is the CIDEr function, and its gradient is approximated as follows:
\begin{equation}
    \nabla_{\theta} L_{R}(\theta) \approx -(r(y_{1:T}^s) - r(y_{1:T}^b)) \nabla_\theta log \  p_\theta(y_{1:T}^s)
    \label{scst_loss}
\end{equation}
where $y_{1:T}^s$ are the sampled captions and $y_{1:T}^b$ are the base predictions. 

\subsection{The End-of-Sequence token in SCST}
 \label{chap_3_problem_description}

Two properties are desirable in an image description: completeness and correctness. 
While the first goal is pursued by the reward maximization, the SCST algorithm provides no explicit control over the latter, which is instead implicitly encouraged by the sequentiality of the decoding process. A token predicted at a specific time step also determines the most likely n-grams in the following ones. Since all n-grams are extracted from linguistically correct references, the final description will be correct, 
at least locally.
Unfortunately, the CIDEr score does not consider a sentence's global correctness, and this aspect can be easily exploited by the SCST if not carefully implemented.
In particular, the algorithm is allowed to produce incomplete descriptions using trivial sentence fragments that almost certainly match some parts of any set of references.
This is the reason why the standard SCST implementation includes the special End-of-Sequence token, abbreviated as \texttt{<Eos>}, in the definition of the n-grams space. With this precaution, the reward function encourages a correct sentence termination leveraging the fact that the \textit{tf-idf} of the \texttt{<Eos>} token out-weights those of function words.

\begin{figure*}[t]
\centering
    \includegraphics[width=0.9\textwidth]{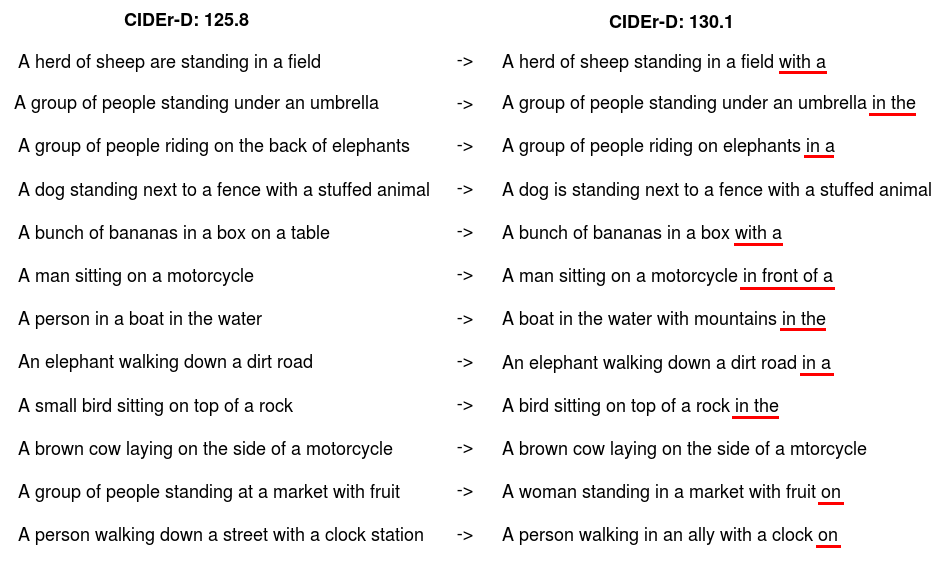}
    \caption{Captions generated by the same model (the Transformer \cite{vaswani2017attention}) trained with different implementations of SCST on the MS-COCO \cite{lin2014microsoft} data set. (Left) The model is optimized by the standard SCST and achieves 125.8 CIDEr-D on the validation set. (Right) The model is optimized by an implementation of SCST in which the \texttt{<Eos>} token is omitted  and achieves 130.1 CIDEr-D on the validation set.}
    \label{img:example_eos_omission}
\end{figure*}

\subsection{The problem of the \texttt{<Eos>} omission} 
\label{impact_eos_in_research}
The inclusion or exclusion of the \texttt{<Eos>} token in the SCST algorithm represents a small and easy-to-overlook detail that significantly impacts a captioning system's performance. In case the \texttt{<Eos>} token is omitted, the descriptions generated by the network are often terminated by trivial sentence fragments such as ``\texttt{and a}",  ``\texttt{in the}", ``\texttt{on top of}" and ``\texttt{in front of}" (more examples in Figure \ref{img:example_eos_omission}).

However, despite the presence of artifacts, they achieve superior performances on popular benchmarks compared to the correct ones (Figure \ref{img:example_eos_omission}).
In particular, the number of additional points yielded by the artifacts can even be greater than the range of values in which different models developed around the same period typically compete.
Therefore, the Image Captioning research field is currently suffering from a lack of transparency and, in some cases lack of awareness over the importance of the \texttt{<Eos>} token in the SCST. The problem can be described from multiple perspectives:
\begin{itemize}
    \item If details over the \texttt{<Eos>} token in the SCST implementation are unavailable, omitted, or simply overlooked, it becomes difficult to compare models in the literature fairly.
    \item Researchers that are aware of the issue are given the difficult choice between less competitive results and poorly formulated outputs.
    \item Finally, researchers that are not aware of the issue (especially the newcomers in the field of Image Captioning) are indirectly encouraged to adopt the implementations that generate compromised sentences because of their superior performances.
\end{itemize}

\section{\texttt{<Eos>} Omission Impact Analysis}\label{sec:ana}
\label{chap_3_quantitative_description}

\subsection{Experimental Setup}

For the qualitative and quantitative analysis of artifacts we implement\footnote{code can be found in \href{https://github.com/jchenghu/captioning_eos}{https://github.com/jchenghu/captioning\_eos}} the Transformer \cite{vaswani2017attention} with 3 layers,  $d_{model}$=$512$ and $d_{ff}$=$2048$, trained
on the COCO 2014 \cite{lin2014microsoft} data set using the Karpathy split \cite{karpathy2015deep}. The Faster-RCNN backbone provided by \cite{anderson2018bottom} is adopted. The learning procedure consists of a first training step on Cross Entropy loss for 8 epochs followed by the CIDEr-D optimization for 20 epochs. The following configurations are adopted:
\begin{enumerate}
    \item batch size of 48, a learning rate of 2e-4 annealed by 0.8 every 2 epochs and warm-up of 10000 in case of Cross Entropy Loss;
    \item batch size of 48, a learning rate of 1e-4 annealed by 0.8 every 2 epochs during the SCST.
\end{enumerate}
Optimization details are provided only for the sake of reproducibility since the artifacts discussed in this work arise regardless of the architecture and optimization details. For the ensemble results, 4 model instances are generated with the aforementioned method differing only in the initialization seed. 
In the experiments, for each seed, the SCST in the Standard and \texttt{No<Eos>} configurations optimize the same pre-trained model.

\subsection{Artifacts Analysis}
\label{section:artifacts_analysis}

The \texttt{<Eos>} token can be omitted in two aspects of SCST:
\begin{enumerate}
    \item during the reward computation;
    \item during the initialization of \textit{tf-idfs};
\end{enumerate}
which leads to 4 implementation instances in case sampled descriptions are tokenized consistently with respect to the ground-truths. Table \ref{tab:eos_impact_table} reports the impact of each configuration over the final descriptions.
\begin{table}[ht!]
  \centering
  \label{tab:1}
  \caption{Impact of the \texttt{<Eos>} token in SCST over the final CIDEr-D score and outputs. ``\textit{tf-idf} Init." refers to the ground truth sentences involved in the calculation of document frequencies, and ``Predictions" refers to the sampled predictions and respective references.}
  \begin{tabular}{| c | c | c |}
 \cline{2-3}
 \multicolumn{1}{c|}{} & {\textit{tf-idf} Init.} & {\textit{tf-idf} Init.} \\
 \multicolumn{1}{c|}{} & w/ \texttt{<Eos>} & w/o \texttt{<Eos>} \\
 \hline
 {Reward} & baseline score & lower score  \\
 w/ \texttt{<Eos>} & no artifacts & with artifacts \\
 \hline
 {Reward} & lower score & higher score \\
 w/o \texttt{<Eos>} & with artifacts & with artifacts \\
 \hline
 \end{tabular}
 \label{tab:eos_impact_table}
\end{table} 
Two cases are the focus of this work since most popular implementations fall into the (\textit{tf-idf} Init. w/ \texttt{<Eos>}, Prediction w/ \texttt{<Eos>}) and  (\textit{tf-idf} Init. w/o \texttt{<Eos>}, Prediction w/o \texttt{<Eos>}) configuration referred as ``Standard" and ``\texttt{No<Eos>}" respectively throughout the rest of this work.

\begin{figure*}[t]
\caption{a) The number of artifacts in the \texttt{No<Eos>} configuration on 5000 test set predictions. b) Average CIDEr-D score of 4 training instances (different seeds) in the Standard and \texttt{No<Eos>} configuration, ``Cleaned" denotes the \texttt{No<Eos>} performance in case artifacts are removed before the evaluation. c) Artifacts distribution. Sequences terminated by ``a"  account for 89.8\% of all cases (top). Histogram of sequences terminated by ``a" (bottom).}
\centering
\includegraphics[width=1.0\textwidth]{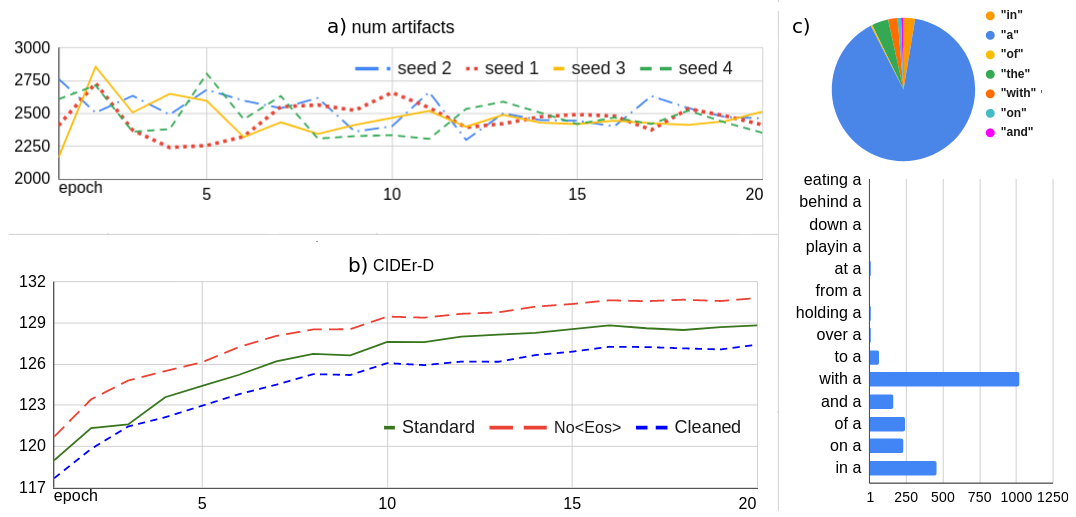}
\label{img:artifacts_images}
\end{figure*}

In the \texttt{No<Eos>} configuration, results are affected by 8 classes of artifacts depending on how sequences are terminated, with the last token belonging to $A$=\{``\texttt{in}", ``\texttt{a}", ``\texttt{of}", ``\texttt{the}", ``\texttt{with}", ``\texttt{on}", ``\texttt{and}"\, ``\texttt{*}"\}, where ``\texttt{*}" represents all the possible remaining cases. While all elements in the set $A$ are just simplifications of longer trivial fragments such as ``\texttt{and a}", ``\texttt{in a}", ``\texttt{with a}" and ``\texttt{in front of}", the case of ``\texttt{on}" may seem acceptable but the token is often part of uncommon formulations such as ``\texttt{a beach with a surfboard on}" and ``\texttt{a street with a bus on}". Nevertheless, ``\texttt{on}" represents only a small fraction of all instances, which mostly end with the ``\texttt{a}" token instead (see Figure  \ref{img:artifacts_images}.c).

Figure \ref{img:artifacts_images}.a showcases the number of artifacts converging to 50\% of the whole testing set as the number of epochs increases. Thus, both correct and compromised sentences are produced by the \texttt{<Eos>} omission, which means the network learns to inject the fragments following a non-trivial and unpredictable criteria for each sequence. 



\begin{table*}[t]
  \centering
  \resizebox{\textwidth}{!}{%
  \caption{Performance comparison the CIDEr-D optimization in Standard and \texttt{No<Eos>} training. ``Cleaned" refers to the \texttt{No<Eos>} results but artifacts are removed prior to the evaluation. $\sum$ refers to the ensemble of the four models and $\varepsilon$ represents the percentage of artifacts. }
  \begin{tabular}{ c | c  c  c  | c  c  c }
 \Xhline{2.2\arrayrulewidth} \bigstrut
 {} & \multicolumn{3}{c|}{Karpathy test split} & \multicolumn{3}{c}{Karpathy validation split} \\
  {} & Standard & \texttt{No<Eos>} ($\varepsilon$) / $\delta$ & Cleaned / $\delta$ & Standard & \texttt{No<Eos>} ($\varepsilon$) / $\delta$ & Cleaned / $\delta$ \bigstrut \\
 \hline
 \bigstrut
 Seed 1 & 128.4 & 131.2 (48.3\%) / +2.8 & 127.8 / -0.6
 & 125.8 & 130.1 (47.5\%) / +4.3 & 126.4 / +0.6 \\
 
 Seed 2 & 129.0 & 130.9 (49.3\%) / +1.9 & 127.4 / -1.6 & 127.0 & 129.9 (48.1\%) / +2.9 & 126.2 / -0.8 \\
 
 Seed 3 & 129.0 & 131.0 (50.3\%) / +2.0 & 127.5 / -1.5 & 127.2 & 129.3 (47.6\%) / +2.1 & 125.7 / -1.5 \\
 
 Seed 4 & 129.1 & 130.7 (50.4\%) / +1.6 & 126.8 / -2.3 & 128.0 & 130.0 (50.6\%) / +2.0 & 126.0 / -2.0 \bigstrut \\
 \hline
 \bigstrut 
 Avg & 128.9 & 130.9 (49.6\%) / +2.0 & 127.3 / -1.1 & 126.9 & 129.8 (48.6\%) / +2.8 & 126.0 / -0.9 \\
$\sum$ & 133.0 & 134.9 (50.2\%) / +1.9 & 131.2 / -1.8 & 131.8 & 133.8 (49.5\%) / +2.0 & 129.8 / -2.0 \bigstrut \\

\Xhline{2.2\arrayrulewidth}
 \end{tabular}
 }
  \label{tab:cider_scores_table}
\end{table*}

Figure \ref{img:artifacts_images}.b and Table \ref{tab:cider_scores_table} showcase that a single model trained with SCST in the \texttt{No<Eos>} configuration consistently outperforms the standard one across all seeds, often by a large margin, with a maximum gain of +2.8 and +4.3 CIDEr-D in the offline test and validation set respectively. Whereas, by removing the artifacts from the latter predictions we observed the opposite trend with a maximum performance decrease of -2.3 and -2.0. Therefore, the increase in score is mostly due to the artifacts and the \texttt{<Eos>} omission poses an obstacle to the generation of semantically meaningful content. Similar behaviour is observed for ensemble performances (referred as $\sum$).

\subsection{Literature classification}
\label{section:revisited}

We sample recent works in the research literature and classify each of them according to the way SCST is implemented. In Section \ref{section:artifacts_analysis} we observed that only half of the evaluated sentences are compromised, which means that if a paper provides only a few correct captioning examples, it is not enough to determine whether the \texttt{<Eos>} token was omitted or not. Because of that, the classification is made through code inspection. The classes and the respective criteria are defined as follows:
\begin{itemize}
    \item Standard: \texttt{<Eos>} token is included in both SCST initialization and reward computation or complete results on either test or validation set are provided;
    \item \texttt{No<Eos>}: \texttt{<Eos>} token is omitted in both initialization and reward computation;
    \item \texttt{Unknown}: the code was not found or it was not available at the time this work was completed. 
\end{itemize}

Table \ref{tab:offline_table_eval} showcases that only 12 of 25 works are confirmed to follow the Standard implementation, 8 fall in the \texttt{No<Eos>} category and 5 are unknown. The State-of-the-art architectures in 2019  \cite{huang2019attention} and 2020 \cite{pan2020x} achieved 129.6 and 131.4 CIDEr-D scores respectively, which showcases the gradual improvement process of the research activity and provides an example of the magnitude of improvements over the years. Unfortunately, such a difference in performance can be lower than the additional score yielded by artifacts (see Section \ref{section:artifacts_analysis}). For instance, if AoANet adopted the \texttt{No<Eos>} configuration, its score would have been comparable to the State-of-the-art performances of the following year (X-Transformer) (see Table \ref{tab:open_source_examples}). 

The amount of \texttt{No<Eos>} implementations in the last years confirms the phenomena described in Section \ref{impact_eos_in_research}.


\begin{table*}[htb!]
  \centering
  \resizebox{\textwidth}{!}{%
  \caption{SCST classification of recent Image Captioning works and their respective performances on the MS-COCO 2014 task. The offline case reports the CIDEr-D score of a single model in contrast to the online evaluation server results where an ensemble is adopted instead with some exceptions denoted with ``\textsuperscript{$\star$}".}
  \begin{tabular}{ c | c | c | c | c | p{7.0cm} }
 \Xhline{2.2\arrayrulewidth}
 \bigstrut
 Year & Work & Offline & Online & SCST & Code inspection \footnote{Prefix https://github.com/} (commit) \bigstrut \\
 \hline
 \bigstrut
  2018 & GCN-LSTM \cite{yao2018exploring} & 127.6 & - & \texttt{Unknown} & Code not found/available \\
 2018 & Up-Down \cite{anderson2018bottom} & 120.1 & 120.5 & Standard & \href{https://github.com/peteanderson80/bottom-up-attention}{peteanderson80/bottom-up-attention} (514e561)  \\
 2019 & HAN \cite{wang2019hierarchical} & 121.7 &  118.2 & \texttt{Unknown} & Code not found/available. \\
 2019 & LBPF \cite{qin2019look} & 127.6 & - & \texttt{Unknown} & Code not found/available \\
 2019 & RDN \cite{ke2019reflective} & 117.3 & 125.2 & \texttt{Unknown} & Code not found/available \\
 2019 & SGAE \cite{yang2019auto} & 127.8 & - & Standard & \href{https://github.com/yangxuntu/SGAE}{yangxuntu/SGAE} (af88115) \\
 2019 & Obj.Rel.Transf. \cite{herdade2019image} & 128.3 & - & Standard &
 \href{https://github.com/yahoo/object\_relation\_transformer}{
 yahoo/object\_relation\_transformer} (6cf5bd8) \\
 2019 & AoANet \cite{huang2019attention}& 129.8 & 129.6 & Standard &
 \href{https://github.com/husthuaan/AoANet}{husthuaan/AoANet} (94ffe17) \\
 2020 & Ruotian Luo \cite{luo2020better} & 129.6 & - & Standard & 
 \href{https://github.com/ruotianluo/self-critical.pytorch}{ruotianluo/self-critical.pytorch} (be1a526) \\
 2020 & M\textsuperscript{2} \cite{cornia2020meshed}& 131.2 & 132.1 & \texttt{No<Eos>} & \href{https://github.com/aimagelab/meshed-memory-transformer}{aimagelab/meshed-memory-transformer} (e0fe3fa) \\
 2020 & X-Transformer \cite{pan2020x} & 132.8 & 133.5 & Standard & \href{https://github.com/JDAI-CV/image-captioning}{JDAI-CV/image-captioning} (d39126d) \\
2020 & Unified VLP \cite{zhou2020unified} & 129.3 & - & Standard & \href{https://github.com/LuoweiZhou/VLP}{LuoweiZhou/VLP} (74c4d85) \\
 2021 & GET \cite{ji2021improving} & 131.6 & 132.5 & \texttt{No<Eos>} & 
 \href{https://github.com/luo3300612/image-captioning-DLCT}{luo3300612/image-captioning-DLCT} (575b4dd) \\
 2021 & DLCT \cite{luo2021dual} & 133.8 & 135.4 & \texttt{No<Eos>} & \href{https://github.com/luo3300612/image-captioning-DLCT}{luo3300612/image-captioning-DLCT} (575b4dd) \\
 2021 & RSTNet \cite{zhang2021rstnet} & 135.6 & 134.0 & \texttt{No<Eos>} & 
 \href{https://github.com/zhangxuying1004/RSTNet}{zhangxuying1004/RSTNet} (e60715f)
 \\ 
 2022 & PureT \cite{wang2022end} & 138.2 &  138.3 & Standard & \href{https://github.com/232525/PureT}{232525/PureT} (8dc9911) \\
 2022 & ExpansionNet \cite{hu2022expansionnet} & 140.4 & 140.8 & Standard & \footnotesize{\href{https://github.com/jchenghu/ExpansionNet_v2}{jchenghu/ExpansionNet\_v2} (365d130)} \\
 2022 & BLIP \cite{li2022blip} & 136.7 & - & Standard & \href{https://github.com/salesforce/BLIP}{salesforce/BLIP} (3a29b74)
 \\
 2022 & CaMEL\cite{barraco2022camel} & 138.9  & 140.0 & \texttt{No<Eos>} & \href{
https://github.com/aimagelab/camel }{aimagelab/camel} (67cb062) \\
 2022 & GRIT \cite{nguyen2022grit} & 144.2 & 143.8 & \texttt{No<Eos>} &  
 \href{https://github.com/davidnvq/grit}{davidnvq/grit} (32afb7e)
 \\
 2022 & S\textsuperscript{2} \cite{ijcai2022p224} & 133.5 & 135.0 & \texttt{No<Eos>} & \href{https://github.com/zchoi/S2-Transformer}{zchoi/S2-Transformer} (c584e4) \\
 2022 & OFA \cite{wang2022ofa} & 154.9 & 149.6\textsuperscript{$\star$} & Standard & \href{https://github.com/OFA-Sys/OFA}{OFA-Sys/OFA} (1809b55) \\
 2022 & ER-SAN \cite{ijcai2022p151} & 135.3 & - & Standard & \href{https://github.com/CrossmodalGroup/ER-SAN}{CrossmodalGroup/ER-SAN} (e80128d) \\
 2022 & CIIC \cite{liu2022show}
  & 133.1 & 129.2\textsuperscript{$\star$}
  & \texttt{Unknown} & Code not found/available \\
2022 & Xmodal-Ctx \cite{kuo2022beyond} & 139.9 & - & \texttt{No<Eos>} & \href{https://github.com/GT-RIPL/Xmodal-Ctx}{GT-RIPL/Xmodal-Ctx} (d927eec) \bigstrut \\
\Xhline{2.2\arrayrulewidth}
 \end{tabular}
 }
  \label{tab:offline_table_eval}
\end{table*} 

\begin{table*}[htb!]
  \centering
  \caption{CIDEr-D performance increase observed in open source projects when the SCST configuration is changed from Standard into \texttt{No<Eos>} mode. Training details can be found in the respective works or repositories.
  }
  \begin{tabular}{ c | c  c  c  c  c  c }
 \Xhline{2.2\arrayrulewidth}
 \bigstrut
 Model & RL epochs & Standard & \texttt{No<Eos>} & Ensemble & Set & $\delta$ \\
 \hline
 \bigstrut
 \multirow{2}{*}{AoANet  \cite{huang2019attention}} & \multirow{2}{*}{25} & 127.6 & 131.0 & \multirow{2}{*}{\xmark} & test & +3.4 \\
   &  & 126.2 & 130.3 &  & val & +4.1 \\
 \hline
 \multirow{2}{*}{X-Transformer \cite{pan2020x}} & \multirow{2}{*}{20} &  131.8 & 133.5 & \multirow{2}{*}{\xmark} & test & +1.7 \\
 &  &  130.1 & 132.3 & & val & +2.2 \\
 \hline
 \multirow{2}{*}{ExpansionNet \cite{hu2022expansionnet}} & \multirow{2}{*}{12} & 143.7 & 145.3 & \multirow{2}{*}{\checkmark} & test & +1.6  \\
  & & 143.0 & 145.7 &  & val & +2.7  \\
 \Xhline{2.2\arrayrulewidth}
 \end{tabular}
 \label{tab:open_source_examples}
\end{table*} 

\section{SacreEOS}\label{sec:sacre}

\subsection{SacreEOS signature}

The lack of transparency and awareness over the \texttt{<Eos>} token in SCST originates from an easy-to-overlook implementation detail. Therefore, the natural solution is to disseminate awareness of the issue. To achieve this goal we introduce SacreEOS, a Python library whose main functionality consists of the generation of signatures that uniquely identify the key aspects of the SCST implementation. In particular, how the \texttt{<Eos>} token is handled. The sharing of the SacreEOS signature accomplishes three objectives:
\begin{enumerate}
    \item it increases transparency and eases the comparison of models;
    \item it informs the reader about the presence or absence of  artifacts (those related to the \texttt{<Eos>} omission) in the results;
    \item last but not least, it spreads awareness of the problem.
\end{enumerate}
We believe this is especially useful in cases of works that do not release the code to the public. 

Established researchers and existing implementations can manually generate the signature using the SacreEOS command line interface. The tool simply asks a few questions regarding the technical aspects of SCST, therefore it does not require any code integration. For new projects instead, SacreEOS consists of an SCST implementation helper, in this case, the signature is provided automatically. Format and signature examples are the following:

Format:
\par\texttt{<scst config>\_<Init>+<metric[args]>+<base[args]>+<Version>}    

Examples:
\par\texttt{STANDARD\_w/oInit+Cider-D[n5,s6.0]+average[nspi5]+1.0.0}\\
\par\texttt{NO<EOS>MODE\_wInit+Cider-D[n4,s6.0]+greedy[nspi5]+1.0.0}\\
\par\texttt{NO<EOS>MODE\_w/oInit+BLEU[n4]+average[nspi5]+1.0.0}

\subsection{Implementation helper and limitations of the approach}

In addition to the functionality of signature generation, the SacreEOS library optionally provides helpful classes to ease the implementation of SCST in future projects. In particular, it covers the following aspects:
\begin{itemize}
    \item \textit{SCST class selection.} Given the number of established works implemented in both Standard and \texttt{No<Eos>} configurations, it is out of the scope of this paper to decide which one is the ``correct" one (the library provides no default option in this regard). However, the tool helps the user to make informed decisions. Classes are currently defined by the reward metric, the reward base and whether the \texttt{<Eos>} token is included or omitted in both initialization and reward computation. 
    \item \textit{SCST initialization.} The library initializes the \textit{tf-idfs} for the reward computation and performs input checks according to the selected class.
    \item \textit{SCST reward computation.} The library currently supports the following reward functions CIDEr, CIDEr-D, CIDEr-R and BLEU. Results are consistent with the official repositories\footnote{
    CIDEr, CIDEr-D, BLEU: \href{https://github.com/vrama91/cider}{github.com/vrama91/cider} \\ CIDEr-R: \href{https://github.com/gabrielsantosrv/coco-caption/tree/master/pycocoevalcap/ciderR}{github.com/gabrielsantosrv/coco-caption}}. Each function is implemented in both Python and C, users can optionally enable the latter version to increase efficiency.    
    \item \textit{Signature generation.} In this case the SacreEOS signature is automatically determined by the class selection and does not require user intervention.
\end{itemize}
The library includes an intricate collection of assertions and input checks on all implementation levels, taylored to each specific class.
Nevertheless, the SacreEOS does not prevent misreporting. In case the signature is manually generated, it relies on the user to provide the correct data.

\section{Related works}
\label{sec:rel}
The work of \cite{rennie2017self} mentioned the role of the End-of-Sequence token. However, it only provided a few qualitative examples and did not report numerical details. Several works in the past focused on improving the evaluation of Image Captioning systems but they mostly proposed alternatives to the CIDEr metric, such as TIGEr \cite{jiang2019tiger}, SPIDEr \cite{liu2017improved}, and CIDEr-R \cite{santos2021cider}. None of them addressed the issue discussed in this work.

The main inspiration of SacreEOS is SacreBLEU \cite{post2018call}, in the field of Machine Translation, where ambiguities can arise from different tokenization and de-tokenization choices that ultimately affect the BLEU score \cite{papineni2002bleu}.

\section{Conclusion}\label{sec:conc}
Our work discussed the role of \texttt{<Eos>} in the Self-Critical Sequence Training and how the lack of transparency and awareness over its function pose an obstacle to the scientific progress in the Image Captioning field. We described the source of the problem from a qualitative and quantitative perspective. We classified recent works in the scientific literature according to the SCST configuration to showcase the pervasiveness and the importance of the matter. Finally, we proposed a possible solution that consists of sharing a unique signature with the help of a Python library called SacreEOS, to enable fair model comparisons and spread awareness regarding the issue. 

\printbibliography[title=Bibliography]

\end{document}